# Fast Optimal Bandwidth Selection for RBF Kernel using Reproducing Kernel Hilbert Space Operators for Kernel Based Classifiers

Bharath Bhushan Damodaran, *Member, IEEE*

*Abstract*—Kernel based methods have shown effective performance in many remote sensing classification tasks. However their performance significantly depend on its hyper-parameters. The conventional technique to estimate the parameter comes with high computational complexity. Thus, the objective of this letter is to propose an fast and efficient method to select the bandwidth parameter of the Gaussian kernel in the kernel based classification methods. The proposed method is developed based on the operators in the reproducing kernel Hilbert space and it is evaluated on Support vector machines and PerTurbo classification method. Experiments conducted with hyperspectral datasets show that our proposed method outperforms the state-of-art method in terms in computational time and classification performance.

## I. INTRODUCTION

Remote sensing imagery provides an important source of information for generating land use/cover maps, monitoring land cover changes, mineral identification, and target detection in military applications [1]. Supervised classification methods are most commonly used to extract the information for the above mentioned applications. The remote sensing images posses unique challenges such as fewer labeled samples in the case of hyperspectral images. Kernel method is considered as the rich set of tools to model various problem (for e.g classification, feature extraction, detection) in remote sensing [1], [2]. These methods implicitly maps the input data into a higher dimensional feature space, where it can be linearly separated. They effectively capture the non-linear relationship present in the data. It is shown that kernel methods is less sensitive to the noise and high dimensionality of the data. Thus, it can effectively deal the curse of dimensionality issue which is more specific to the hyperspectral image.

Due to these peculiarities, kernel based solutions have been extensively developed to solve classification tasks (SVM, PerTurbo) [2]–[4], feature selection task (SVM based feature selection, Sparse HSIC) [5], [6], feature extraction (kernel LDA, kernel ICA) [7], and target detection tasks. These methods had reported state of the art results. The kernel methods are still relevant in the deep learning era, especially when the labeled samples are limited, and also with recent invent of Random Fourier features [8]. Due to the Random Fourier features, kernel methods are no more limited to small scale datasets, and they also has shown outstanding performance in the classification and regression tasks [8]–[11]. The key to the performance of kernel methods is the choice of the kernel and its parameters. Especially, Gaussian radial basis (RBF) kernel has shown to be an optimal choice of kernel for variety of tasks. The significance of the Gaussian kernel depends on its bandwidth parameter. Thus, the improper selection of the Gaussian bandwidth parameter leads to the inferior results for the classification tasks.

In literature, mostly the bandwidth parameter of RBF kernel is selected based on traditional grid search method and $k$-fold cross validation technique with respect to the performance of the underlying kernel based classifier [2], [5], [12]. This approach works quite well in practice, however the computational complexity is much higher due to the evaluation of each bandwidth parameter with respect to the $k$-fold classification task. Recently in [13], [14] a method is proposed to estimate the bandwidth parameter automatically. They assume that the RBF kernel with optimal bandwidth parameter will provide better class separability in the reproducing kernel Hilbert space (RKHS), and the solution is obtained through optimization procedure. They showed that their method was able to reduce the computational time compared to grid search and $k$-fold cross validation method, without degrading the classification performance. However, it seems from our experimental evaluation that the computational complexity is still higher, because it might take several iterations to converge to the solution of the optimization problem. Furthermore, the solution and convergence also might depend on the initialization to the optimization problem. Due to this limitation, we further attempt to reduce the computational time to find

The author is with Univ.Bretagne-sud, IRISA, Vannes, France. Corresponding author: Bharath Bhushan Damodaran (email: bharath-bhushan.damodaran@irisa.fr).

the optimal bandwidth parameter of the RBF kernel. In this letter, we propose a fast and efficient bandwidth selection method based on the RKHS operators. The proposed method is evaluated on the two datasets, and we have shown that our method not only decreases the computational time significantly, but also increases the classification accuracy.

The rest of the paper is organized as follows: Section II describes the concepts related to our proposed method, Section III presents our proposed method, experimental results are presented in Section IV and finally conclusions and future perspectives are derived in Section V.

## II. KERNELS AND INDEPENDENCE MEASURES

In this section, we describe the Gaussian kernel and Hilbert Schmidt operators in the RKHS which are core to our proposed method.

### A. Gaussian Kernel

Let $S = \{(\mathbf{x_1}, y_1), (\mathbf{x_2}, y_2), \ldots, (\mathbf{x_n}, y_n)\} \in \mathbb{R}^d \times \{\omega_1, \ldots, \omega_L\}$ be the training set with $N$ pairs of training samples $\mathbf{x_i}$ and their corresponding class labels $y_i$. The similarity between the training data points can be represented by the Gram matrix in the RKHS as follows:

$$\begin{aligned} \boldsymbol{K_{ij}} &= k(\mathbf{x_i}, \mathbf{x_j}) = \phi(\mathbf{x_i})^{\mathbf{T}}.\phi(\mathbf{x_j}) \\ &= \exp\left(-\gamma \|\mathbf{x_i} - \mathbf{x_j}\|^{\mathbf{2}}\right), \end{aligned} \quad (1)$$

where $k(.,.)$ is a Gaussian radial basis kernel function, $\mathbf{x_i}$ and $\mathbf{x_j} \in S$, $\phi$ is the mapping from the original space into RKHS space, and $\gamma$ is the bandwidth parameter of the Gaussian RBF kernel. The performance of the kernel based methods (for e.g, support vector machines) depends on appropriate choice of the bandwidth parameter of the RBF kernel. This letter proposes a fast and effective method to estimate the bandwidth parameter of the RBF kernel. In the next subsections, we describe the necessary theoretical details for our proposed method.

### B. Kernel Independence Measures

*1) Hilbert Schmidt Independence Criteria:* Hilbert Schmidt Independence Criteria (HSIC) compares the geometry of kernel embeddings by the use of a statistical independence criterion. It measures the independence between two sets of random variables [15]. Let $X$ and $Z$ be the two random variables, from which the samples $(\mathbf{x}, \mathbf{z})$ can be drawn from the probability density function of $X$ and $Z$. The non-linear mapping function is defined on each element of $X$, as $\phi(\mathbf{x}) \in F$ from $\mathbf{x} \in X$ to the feature space $F$, such that the inner product between the features is given by a kernel function $k(\mathbf{x}, \mathbf{x}') = \langle \phi(\mathbf{x}), \phi(\mathbf{x}') \rangle$, and $F$ is the associated reproducing kernel Hilbert space. In a similar manner, let $G$ be the RKHS on $Z$ with kernel $l(.,.)$ and the mapping function $\psi(\mathbf{z})$. Then the cross-covariance operator between these two mapping functions can be defined as a linear operator $C_{xz}: G \to F$, such that

$$\begin{aligned} C_{\mathbf{xz}} &= E_{\mathbf{xz}}\left[(\phi(\mathbf{x}) - \mu_{\mathbf{x}}) \otimes (\psi(\mathbf{z}) - \mu_{\mathbf{z}})\right] \\ &\Rightarrow E_{\mathbf{xz}}\left[\phi(\mathbf{x}) \otimes \psi(\mathbf{z})\right] - \mu_{\mathbf{x}} \otimes \mu_{\mathbf{z}}, \end{aligned} \quad (2)$$

where $\otimes$ is a tensor product. The HSIC is defined as the squared Hilbert Schmidt norm of Eq. (2). It has been shown that the HSIC can be expressed in terms of kernel [15], and the empirical estimate of the HSIC is given as follows:

$$HSIC(\boldsymbol{Z}, \boldsymbol{F}, \boldsymbol{G}) = (m-1)^{-2} tr(\boldsymbol{KCLC}), \quad (3)$$

where $m$ is the number of observations (samples), $\boldsymbol{C}, \boldsymbol{K}, \boldsymbol{L} \in \mathbb{R}^{m \times m}$, $\boldsymbol{K}_{ij} = k(\mathbf{x_i}, \mathbf{x_j})$, $\boldsymbol{L}_{ij} = l(\mathbf{z_i}, \mathbf{z_j})$, $\boldsymbol{C_{ij}} = \delta_{ij} - m^{-1}$, ($\delta_{ij} = 1$ if $i = j$, zero otherwise) is the centering matrix, and $tr$ is the trace operator. The independence between the two random variables in the Hilbert space can be obtained by Eq. (3).

*2) Kernel Target Alignment:* The kernel target alignment (KTA) method is an another way to measure or compare the similarity between the two kernels [16]. For any two kernels $\boldsymbol{K}$ and $\boldsymbol{L}$, the degree of kernel alignment is defined as

$$KTA(\boldsymbol{Z}, \boldsymbol{F}, \boldsymbol{G}) = \frac{tr(\boldsymbol{KL})}{\sqrt{tr(\boldsymbol{KK}) tr(\boldsymbol{LL})}}. \quad (4)$$

This measure can be seen as the cosine angle between the two vectors. Based on these two measures in the next section we describe the proposed method.

## III. PROPOSED BANDWIDTH SELECTION METHOD

The proposed bandwidth selection method for RBF kernel is developed based on the kernel independence measures. The kernel similarity measures HISC and KTA attain high value when the two kernels have similar structure and low value when they are dissimilar. We define two kernels, the first one is constructed from the training data as shown in eq. (1), i.e,

$$K(\mathbf{x_i}, \mathbf{x_j}, \gamma) = \exp\left(-\gamma \|\mathbf{x_i} - \mathbf{x_j}\|^{\mathbf{2}}\right), \quad (5)$$

and the second kernel is constructed from the labels of the training data, which is called as ideal kernel or target kernel. We use the delta kernel to construct the ideal kernel [6] and it is defined as

$$L(y, y') = \begin{cases} \frac{1}{m_y} & \text{if } y = y', \\ 0 & \text{otherwise}, \end{cases} \quad (6)$$

where $m_y$ is the number of samples in class $y$. The non-linear mapping of the data points is well clustered in the RKHS, if the structure of the kernel defined in eq. (5) is similar to the ideal kernel. The choice of the bandwidth parameter ($\gamma$) determines the structure of the kernel matrix, thus being a crucial parameter to be tuned. As the ideal kernel or target kernel is the desired structure, we seek for the $\gamma$ for which eq.5 has similar structure to the ideal kernel.

Thus, the objective of our work can be formulated as follows

$$\gamma^* = \max_{\gamma} \text{KID}\left(L, K\left(\gamma\right)\right), \quad (7)$$

where $\gamma^*$ is the optimal choice of bandwidth parameter, KID is the kernel independence measure using either HSIC or KTA. The classification procedure using our proposed method is shown in algorithm 1. The procedure is similar to grid search method, however instead of performing $k$-fold classification for each $\gamma$, we just measure the kernel independence measure using eq. (5) and (6).

---

**Algorithm 1** Proposed bandwidth selection method
---

1: **procedure** INPUT DATA(Data)
**Require:**
   X: Training samples, y: corresponding labels
2:   $\gamma = 2^\beta, \beta = \{-15, -13, \ldots, 4, 5\}$
3:   **for** each $\gamma$ **do**
4:     Compute the K and L from eq. (5) and (6)
5:     $m_i = \text{KID}\left(L, K\left(\gamma\right)\right)$
6:   **end for**
7:   $index = \arg\max m$, $\gamma^* = \gamma(index)$
8:   Cross validate for the other parameters of classifier
9:   Train the classifier
10:  Predict the samples using trained model
11: **end procedure**

---

IV. EXPERIMENTAL RESULTS AND DISCUSSION

*A. Datasets*

In order to evaluate the proposed bandwidth selection method, experiments have been conducted with real world hyperspectral datasets. A detailed description of these data is provided below.

*1) Pavia University:* The first hyperspectral data considered here was collected over the University of Pavia, Italy by the ROSIS airborne hyperspectral sensor in the framework of the HySens project managed by DLR (German national aerospace agency). The ROSIS sensor collects images in 115 spectral bands in the spectral range from 0.43 to 0.86 $\mu$m with a spatial resolution of 1.3 m/pixel. After the removal of noisy bands, 103 bands were selected for experiments. This data contains 610 $\times$ 340 pixels with nine classes of interest. The training and testing samples are provided along with data and are used to perform quantitative evaluation (see Tab. I).

TABLE I: Pavia University Training and Testing Samples.

| No | Class name | Training | Testing |
|---|---|---|---|
| 1 | Asphalt | 548 | 6641 |
| 2 | Meadows | 540 | 18649 |
| 3 | Gravel | 392 | 2099 |
| 4 | Trees | 524 | 3064 |
| 5 | Metal Sheets | 265 | 1345 |
| 6 | Soil | 532 | 5029 |
| 7 | Bitumen | 375 | 1330 |
| 8 | Bricks | 514 | 3682 |
| 9 | Shadows | 231 | 947 |
| | Total | 3921 | 42776 |

*2) City of Pavia:* The second hyperspectral data was collected over the City of Pavia, Italy by the ROSIS sensor. The spectral and spatial resolution configurations are similar to Pavia University dataset. After the removal of noisy bands, 102 bands were selected for experiments. This dataset contains 1096 $\times$ 715 pixels with nine classes of interest. Similarly to Pavia University, the City of Pavia dataset comes with training and testing samples (see Tab. II).

TABLE II: City of Pavia Training and Testing Samples

| No | Class name | Training | Testing |
|---|---|---|---|
| 1 | Water | 824 | 65971 |
| 2 | Trees | 820 | 7598 |
| 3 | Gravel | 824 | 3090 |
| 4 | Trees | 808 | 2685 |
| 5 | Metal Sheets | 820 | 6584 |
| 6 | Soil | 816 | 9248 |
| 7 | Bitumen | 808 | 7287 |
| 8 | Bricks | 1260 | 42826 |
| 9 | Shadows | 476 | 2863 |
| | Total | 7456 | 148152 |

*B. Experimental design*

The effectiveness of the proposed bandwidth selection method for the Gaussian RBF kernel is assessed on the two kernel based classification methods, support vector machines (SVM) [2], and PerTurbo [3]. The choice is due to their demonstrated performance in the classification tasks [2], [5]. The proposed method has two variants depending on the kernel independence measure used and we label them as KID-HSIC and KID-KTA. The performance of our proposed method is compared with the conventional grid search and cross validation method and recently proposed method [13], we label them as $k$-fold cross validation ($k$-fold CV), and class separability (CS) method.

The bandwidth parameter of the RBF kernel are varied from $\gamma = 2^\beta, \beta = \{-15, -13, \ldots, 4, 5\}$. For our proposed method, once we obtained optimal bandwith parameter ($\gamma^*$), the cost function $C = 2^\alpha$, $\alpha = \{-5, -4, \ldots, 15\}$ of the SVM classifier and regularization parameter $\tau = \{10^{-6}, 10^{-5}, \ldots, 0.99\}$ of the PerTubro classifier are tuned using five fold classification. For the conventional method ($k$-fold CV), the parameters of the SVM and PerTurbo classifiers were automatically tuned with the grid search method using five fold classification. Lastly, the class separability based method (CS) determines the bandwidth parameter based on minimizing the optimization problem [13], and the cost function of the SVM and regularization parameter of PerTurbo classifier is tuned based on grid search and five fold cross-validation method. For more details about determining bandwidth parameter based on CS method, please refer to [13], [14].

The effectiveness of the proposed method is assessed in terms of computational efficiency, overall accuracy (OA), and kappa coefficient (KC) with respect to the considered classifiers. The experiments are conducted in two settings. Firstly, the bandwidth parameter selection is performed on the original training data, and secondly, the impact of the proposed and state-of-the-art methods were assessed on the different number of training samples.

*C. Experimental results*

TABLE III: Experimental results of our proposed method and the existing methods with SVM classifier for Pavia University dataset. $\gamma^*$ denotes the selected (estimated) bandwith parameter of the Gaussian RBF kernel, $C^*$ denotes the selected cost function of the SVM classifier, OA and KC indicates the overall accuracy in (%) and the kappa coefficient respectively. Values in **bold**, and ***italics bold*** indicates the best method with respect to OA (KC), and computational time respectively.

| Methods | $\gamma^*$ | $C^*$ | CV time (in sec) | OA (in %) | KC |
|---|---|---|---|---|---|
| $k$-fold CV | 0.0078 | 128 | 1326 | 80.28 | 0.7542 |
| CS | 0.0131 | 128 | 523 | 80.94 | 0.7617 |
| KID-KTA | 0.0313 | 128 | ***32*** | 80.21 | 0.7528 |
| KID-HSIC | 0.0156 | 128 | 129 | **81.03** | **0.7623** |

*1) Pavia University:* Table III reports the experimental results of the proposed method and state of the art approaches with the SVM classifier for Pavia University dataset. Firstly, our proposed method did not deteriorate the classification accuracy, on the otherhand it slightly outperformed the existing methods, which is evident from the classification accuracy and kappa coefficient. This is an interesting observation because our method did not select the optimal bandwidth parameter based on the classification accuracy, unlike the $k$-fold CV approach. Among our proposed method, KID-HSIC slightly outperforms the KID-KTA, this is inline with the knowledge that HSIC is the better measure to compare the similarity between the probability distributions in the Hilbert space.

Apart from the classification point of view, our proposed method reduces the computational time complexity more significantly than the conventional $k$-fold CV and class separability (CS) based approaches. The CV time mentioned in the tables is the total time required to compute all the parameters of the classifier (for e.g $\gamma^*$, and $C^*$ for the SVM classifier). When the KID-KTA is considered, we are $41X$, $16X$ times faster than $k$-fold CV and CS method. This is a significant reduction in time compared to the earlier approaches. KID-HSIC is $4X$ times slower than the KID-KTA. The results with PerTurbo classifier is presented in table IV, and shows that our proposed method not only decreased the computational time, but also increased the classification accuracy upto 3% compared to $k$-fold CV approach. This shows our proposed method explores the parameter space better than existing methods.

Furthermore, from the tables III, IV it is interesting to observe that the selected optimal $\gamma^*$ value is different with different methods. This shows that these methods operate in a complimentary way while estimating the optimal bandwidth parameter. However, it can be concluded that they lie in the nearby regions of the parameter space. Thus leading to the different classification accuracies, but not drastically different.

TABLE IV: Experimental results of our proposed method and existing methods with the PerTubro classifier for Pavia University dataset. $\tau^*$ denotes the regularization parameter of the PerTurbo classifier.

| Methods | $\gamma^*$ | $\tau^*$ | CV time (in sec) | OA (%) | KC |
|---|---|---|---|---|---|
| $k$-fold CV | 0.125 | 0.1 | 277 | 0.7543 | 0.6938 |
| CS | 0.0131 | 0.01 | 505 | 0.7796 | 0.7245 |
| KID-KTA | 0.0313 | 0.1 | ***13*** | 78.14 | 0.726 |
| KID-HSIC | 0.0156 | 0.01 | 110 | **78.29** | **0.7283** |

*2) Pavia Centre:* Tables V and VI reports the experimental results with the SVM and PerTurbo classifier for Pavia Centre dataset. The classification accuracies and kappa coefficient of the proposed methods are similar to the existing methods. The computational complexity of the $k$-fold CV is worse, where as our proposed method (KID-KTA) is $64X$ times faster than the $k$-fold CV approach. Furthermore, our method is also better than CS method. This highlights that our method scales well with the large number of training samples compared to the existing methods. The higher values of computational time for the Pavia centre data set is due

TABLE V: Experimental results of our proposed method and existing methods with the SVM classifier for Pavia Centre dataset.

| Methods | $\gamma^*$ | $C^*$ | CV time (in sec) | OA (%) | KC |
|---|---|---|---|---|---|
| $k$-fold CV | 0.0313 | 32 | 4094 | 97.04 | 0.9584 |
| CS | 0.0118 | 128 | 1570 | **97.77** | **0.9686** |
| KID-KTA | 0.0625 | 128 | *64* | 96.71 | 0.9538 |
| KID-HSIC | 0.0156 | 128 | 600 | 97.47 | 0.9644 |

to the large number of training samples compared to the Pavia University dataset.

TABLE VI: Experimental results of our proposed method and existing methods with the PerTurbo classifier for Pavia Centre dataset.

| Methods | $\gamma^*$ | $\tau^*$ | CV time (in sec) | OA (%) | KC |
|---|---|---|---|---|---|
| $k$-fold CV | 0.5 | 0.2 | 1397 | 96.52 | 0.951 |
| CS | 0.0118 | 0.01 | 1563 | **97.07** | **0.9588** |
| KID-KTA | 0.0625 | 0.1 | *50* | 96.85 | 0.9556 |
| KID-HSIC | 0.0156 | 0.01 | 595 | 97.04 | 0.9583 |

*3) Impact with size of the samples:* Here, we investigate the impact of the proposed method and existing methods with different number of sample sizes, especially with large sample sizes. As we have demonstrated the potential of our method with two datasets and two classifiers, here we consider only one dataset (Pavia University) and one classifier (SVM). From the available samples, we randomly choose different set of samples as training samples and remaining samples are used as testing samples. Figure 1 shows the computational time of different methods to estimate the parameters of the SVM classifier. From the Fig. 1 we can infer that the $k$-fold CV method has the quadratic time complexity as the sample size increases, thus time consuming when the samples are large. When the CS method is considered, the computational time reduced significantly compared to $k$-fold CV, however still the CV time is large when the sample size increases. On contrary to this, both of our proposed methods are not much sensitive to the large number of sample sizes and the reduction in the computational time is very drastic. Further, we show that classification accuracy of the SVM classifier using our proposed method is similar to existing methods (see Fig. 2). This highlights that our proposed methods has an advantage of reducing the computational time significantly without degrading the classification accuracy. Thus, can be considered as an alternative choice to efficiently estimate the bandwidth parameter of RBF kernel in practical applications.

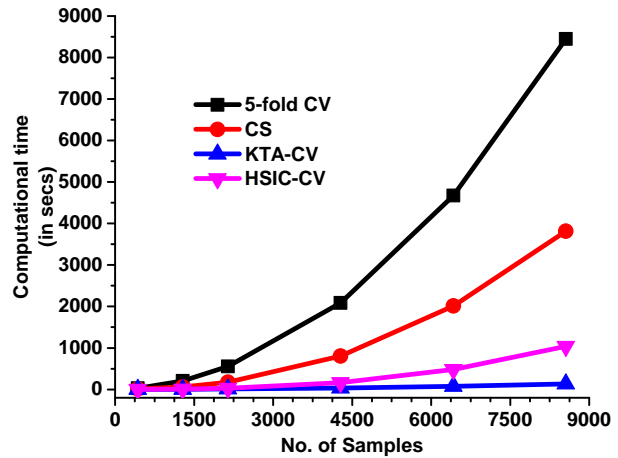

Fig. 1: Computational time (in secs) of the different methods with number of training samples for the SVM classifier with Pavia University dataset.

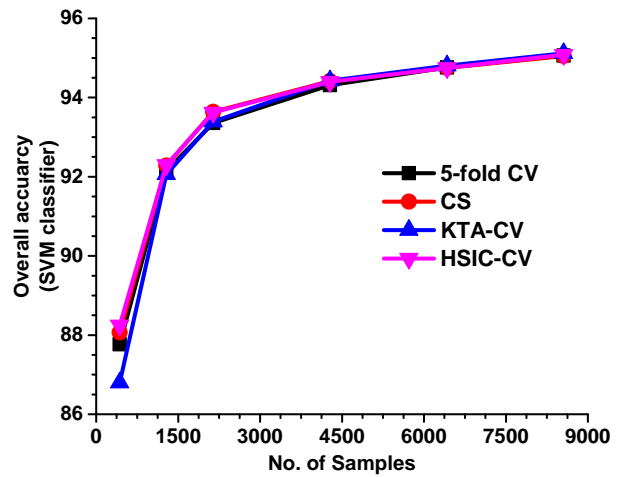

Fig. 2: Classification accuracy (in %) of the different methods with different number of training samples for the SVM classifier with Pavia University dataset.

## V. CONCLUSION

In this letter, we proposed an fast and accurate method to estimate the bandwith parameter of the Gaussian RBF kernel. The proposed methods are developed by measuring the similarities between the distributions in the RKHS using kernel independence measures. Experimental results showed that our proposed method reduced computational time drastically compared to the state-of-the-art methods. Furthermore we also showed that our method resulted with similar or better classification accuracy than state-of-the-art methods. Thus, our method can be considered as an alternative choice to efficiently estimate the bandwidth parameter of RBF kernel in practical applications. Future work will be in directions of exploring further to scale for very large sample size with the usage of random Fourier features in the kernel independence measure [8], [17].

ACKNOWLEDGMENT

This work was supported in part by the French Agence Nationale de la Recherche under Project ANR-13-JS02-0005-01 (Asterix project) and in part by the People Programme (Marie Curie Actions) of the European Unions Seventh Framework Programme (FP7/2007-2013) under REA Grant PCOFUND-GA-2013-609102, through the PRESTIGE programme coordinated by Campus France. The authors would like to thank Prof. P. Gamba for providing ROSIS hyperspectral images.